\documentclass{article}

\usepackage[nonatbib,preprint]{neurips_2021}

\usepackage[utf8]{inputenc} % allow utf-8 input
\usepackage[T1]{fontenc}    % use 8-bit T1 fonts
\usepackage{hyperref}       % hyperlinks
\usepackage{url}            % simple URL typesetting
\usepackage{booktabs}       % professional-quality tables
\usepackage{amsfonts}       % blackboard math symbols
\usepackage{nicefrac}       % compact symbols for 1/2, etc.
\usepackage{microtype}      % microtypography
\usepackage{xcolor}         % colors
\usepackage{graphicx}
\usepackage{lineno,amsfonts,amsmath}

\usepackage{algpseudocode,algorithm}

\algrenewcommand\algorithmicrequire{\textbf{Precondition:}}  
\algrenewcommand\algorithmicensure{\textbf{Postcondition:}}

\title{Sequential memory improves sample and memory efficiency in Episodic Control}
\author{
  Ismael T. Freire \\
  Donders Institute for Brain, Cognition and Behaviour - Centre for Neuroscience (DCN-FNWI) \\
  Radboud University, Nijmegen, the Netherlands \\
  \texttt{ismael.freire@donders.ru.nl} \\
  \And
  Adrián F. Amil \\
  Donders Institute for Brain, Cognition and Behaviour - Centre for Neuroscience (DCN-FNWI) \\
  Radboud University, Nijmegen, the Netherlands \\
  \texttt{adrian.fernandezamil@donders.ru.nl} \\
   \And
  Paul F.M.J. Verschure \\
  Donders Institute for Brain, Cognition and Behaviour - Centre for Neuroscience (DCN-FNWI) \\
  Radboud University, Nijmegen, the Netherlands \\
  \texttt{paul.verschure@donders.ru.nl} \\
}

\begin{document}

\maketitle

\begin{abstract}
Deep reinforcement learning (DRL) algorithms are known for their sample inefficiency, requiring extensive episodes to reach optimal performance. Episodic Reinforcement Learning (ERL) algorithms aim to overcome this issue by using extended memory systems to leverage past experiences. However, these memory augmentations are often used as mere buffers, from which isolated events are resampled for offline learning (e.g., replay). In this paper, we introduce Sequential Episodic Control (SEC), a hippocampal-inspired model that stores entire event sequences in their temporal order and employs a sequential bias in their retrieval to guide actions. We evaluate SEC across various benchmarks from the Animal-AI testbed, demonstrating its superior performance and sample efficiency compared to several state-of-the-art models, including Model-Free Episodic Control (MFEC), Deep Q-Network (DQN), and Episodic Reinforcement Learning with Associative Memory (ERLAM). Our experiments show that SEC achieves higher rewards and faster policy convergence in tasks requiring memory and decision-making. Additionally, we investigate the effects of memory constraints and forgetting mechanisms, revealing that prioritized forgetting enhances both performance and policy stability. Further, ablation studies underscore the critical role of the sequential memory component in SEC. Finally, we propose a novel perspective on how fast, sequential hippocampal-like episodic memory systems could support both habit formation and deliberation in artificial and biological systems.

\end{abstract}

\section{Introduction}
The increasing popularity of Deep Reinforcement Learning (DRL) has been driven in recent years by its ability to reach human-level performance in several domains that were traditionally considered hallmarks of human intelligence. For instance, beating world champions in games like chess and Go  \cite{Silver2018}, or more recently, in complex real-time multiplayer games like Starcraft \cite{vinyals2019grandmaster} and DOTA \cite{berner2019dota}. However, in order to achieve such remarkable feats, these algorithms require orders of magnitude more data to learn from than humans \cite{lake2017building}. This is an illustration of the difference in the sampling efficiency of DRL and the human brain.

The sample inefficiency problem in DRL refers to the large amounts of data that these methods require to reach a human level of performance \cite{Marcus2018-deep}. In the case of the classic Atari games, DRL systems require millions of samples to reach human-level results \cite{Mnih2015}. In recent scenarios involving more complex tasks, the amounts of samples can reach up to several billion \cite{baker2019emergent}. A number of sources of inefficiency and slowness in DRL have been already identified  \cite{Botvinick2019}. One pertains to an intrinsic feature of such systems: the gradient-based updates that slowly drive the learning of both policy and value functions limit the speed at which these systems can achieve optimal performance. 

Recently, variants of Episodic Reinforcement Learning (ERL) algorithms have been proposed to overcome the intrinsic limitations of the gradient-based methods of DRL. In particular, the introduction of a memory system that allows the algorithm to make use of previously successful experiences to speed up the learning of an optimal policy. Within ERL, two main approaches have been followed. First, enhancing and bootstrapping the offline learning capacity of a Deep Q-Network (DQN) by replaying past experiences stored in a memory buffer \cite{Hansen2018,zhu_episodic_2020,Lin2018,Lee2018}. Second, using the stored events in memory for direct control by generating the policy directly from the memory buffer \cite{Blundell2016,Pritzel2017,Yalnizyan2021forgetting}. Both of these solutions are partially inspired by some features of the hippocampal episodic memory in biological systems, in particular, the known phenomenon of the acquisition, retention, and replay of stored behavioral sequences \cite{davidson2009hippocampal}. 

Although capable of improving the sample efficiency of DRL, ERL methods also face a problem of memory efficiency. The efficient use of a limited memory buffer capacity is especially critical when embedding such algorithms in embodied systems such as robots that face strict real-time computational and storage limitations \cite{Khamassi2020adaptive} - a problem that the brain also has to deal with \cite{Lisman1995storage,Jensen2001dual}. However, studying the effects of incorporating memory constraints in episodic reinforcement learning models is something that previous approaches have not considered \cite{Ramani2019short}, with the recent exception of \cite{Yalnizyan2021forgetting} which studied the role of forgetting in episodic control agents in 2D discrete grid-world settings. 

Moreover, thus far Episodic Reinforcement Learning models do not capitalize on the sequential nature of the temporal unfolding of events that are reflected in the structure of hippocampal memory \cite{Buzsaki2018space, Lisman2009prediction, Verschure2014}. Notably, states in ERL are treated as being independent samples of a distribution of world states \cite{Blundell2016,Pritzel2017,Yalnizyan2021forgetting}, instead of being experienced, stored, and retrieved in a sequential manner due to the embodiment of the agent and the continuity of real-world interaction \cite{Buzsaki2018space,merleau1964primacy}. Indeed, the idea that integrated episodes may guide behavior, echoing the efficiency of human episodic memory, has been supported by neuroscientific evidence \cite{bornstein2017reinstated, wimmer2012preference, wu2021inference}. These studies suggest that behavior is informed by compression of sensory data into integrated episodes, reflecting trajectories through environmental and cognitive spaces, rather than isolated state-action pairs. The implications of treating experiences as integrated sequences have been discussed in the literature, emphasizing the importance of sequentiality in both animal and human cognition \cite{johnson2007neural, ludvig2015priming, wang2022mixing}.

Based on this evidence, we propose that the performance of ERL methods can be further improved by incorporating additional aspects of mammalian episodic memory \cite{Lisman2009prediction, Gershman2017, santos2021epistemic}. In particular, we suggest that the hippocampus memory system implements core features that solve the sampling efficiency challenge. Notably, the hippocampus has been argued to compress sensory data to generate efficient representations of the world's state \cite{santos2021epistemic}. Beyond this autoencoder-like function that explains many features of hippocampal dynamics \cite{santos2021entorhinal, amil2024discretization}, we also consider the conjunctive nature of event representations \cite{renno2010mechanism}, and their sequential scaffolding and chaining into coherent, repeatable and goal-oriented memory episodes \cite{Buzsaki2018space,estefan2019coordinated}. This sequential linking of events is in turn enabled by a particular winner-take-all selection mechanism \cite{De_Almeida2009-emax} that filters the unfolding sequence of behavioral events \cite{skaggs1996phaseprecession}. These sequences are time-multiplexed at faster neuronal timescales and maintained in short-term memory via a nested-frequencies code \cite{Lisman1995storage}, which allows for fast long-term consolidation and retrieval and serves key cognitive functions such as mind-travel at decision points \cite{redish2016vicarious} and sequential decision-making \cite{Clayton1998episodic, Foster2012sequence}. Further, empirical evidence points to the prospective replay of rewarded sequences \cite{Mattar2018} playing a key role in bootstrapping learning and biasing decision-making towards previously successful actions \cite{eichenbaum2017memory, estefan2019coordinated} while they are coordinated by the active pursuit of goals by the agent \cite{estefan2021volitional}. Here we incorporate these features to enhance episodic control of reinforcement learning agents, focusing on the sequential binding of events into goal-oriented episodes and the way the retrieval of those affects the learning dynamics.

In summary, in this paper, we explore the role of sequentiality, memory capacity, and forgetting in Episodic Reinforcement Learning. We present a novel algorithm, Sequential Episodic Control, that leverages the sequential nature of the stored experiences for direct control. We demonstrate how imposing a sequential inductive bias on the retrieval of memories for action selection favors both overall performance and sample efficiency in several naturalistic foraging benchmarks. Moreover, we also show how this sequential inductive bias enhances memory efficiency by an order of magnitude compared to the same episodic control model without such a feature. Finally, we observe that forgetting slightly enhances both efficiency and policy stability, but to a lesser degree than the introduction of the sequential bias.

\section{Background and related work}
\label{sec:related_work}
Following Thorndike's law of effect, the goal of a reinforcement learning agent is to maximize reward through its interaction with the environment \cite{Sutton2018}. This is usually operationalized as maximizing the expected discounted return $ R_t=\sum_{k=0}^{T}{\gamma^kr_{t+k}} $, where T is the length of an episode and $ \gamma\ \epsilon\ (0,1] $ is the discount factor. Given state $ s_t\epsilon\ S $ of the environment, the agent takes action $ a_t\epsilon\ A $ following its policy $ \pi(s_{t,\ }a_t) $ which brings about a reward $ r_{t+1}\epsilon\ R $ leading to a novel state of the environment $ s_{t+1} $. In Q-learning \cite{watkins1992q}, the agent learns the action-value function $ Q^\pi\left(s,\ a\right)=\mathbb{E}[R_t|s_t=s,a] $ by computing the expected rewards obtained by acting on a given state. In Deep Reinforcement Learning methods, such as DQN \cite{Mnih2015}, this function is parametrized by a deep neural network to approximate the optimal action-value function. 

Most of the work on Episodic Reinforcement Learning has tried to improve the sample efficiency of parametric models like DQN, either by bootstrapping their learning through an extended memory system or by value propagation methods that capitalize more on the experiences that yielded higher rewards in the past. The so-called Episodic Memory Deep Q-Network (EMDQN) adds a memory buffer parallel to a Deep Q-Network and shows that having coordinated systems leads to faster reward propagation and higher sample efficiency as compared to the standard Deep Q-Network \cite{Lin2018}. In contrast, the Episodic Backward Update (EBU) model propagates the value of a state to its previous states after sampling a complete episode \cite{Lee2018}. This modification allows the EBU model to achieve the same mean human normalized performance on several Atari benchmarks as DQN while using only 5\% of the data. An alternative model called Episodic Reinforcement Learning with Associative Memory (ERLAM) stores the different behavioural trajectories in a graph instead of a dictionary and associates different nodes to create an instance-based reasoning model that is more sample efficient \cite{zhu_episodic_2020}. Indeed, this follows an earlier proposal by Kubie on the graph-like features of hippocampal memory \cite{kubie2009heading} which has been successfully translated to robot models of foraging \cite{mathews2009insect}.

In contrast to ERL approaches that capitalize on different forms of episodic memory to accelerate learning, episodic control models, such as Model-Free Episodic Control (MFEC), remove all gradient-based methods using a non-parametric instance-based way of learning \cite{Blundell2016}. MFEC records rewarding experiences in a tabular memory and follows a policy that capitalizes on those stored events. MFEC updates its action-value estimates by storing in memory the highest Q values experienced in a state. When encountering a state, the system consults this memory and picks the state-action pair that gave the highest reward. When faced with novel states, their value is approximated by averaging the action-values of the $k$ state neighbours, using Euclidean distance as a similarity metric. Neural Episodic Control builds upon MFEC by adding a so-called, differentiable neural dictionary (DND) which stores slow-changing state representations and fast-updating value estimates, retrieving those values for efficient action selection by using context-based lookup \cite{Pritzel2017}.

The Sequential Episodic Control (SEC) model that we present in this paper departs from the previous literature on episodic control in several aspects. First, it considers state-action pairs as integrated representational primitives which reflects hippocampal coding \cite{renno2010mechanism}. Second, it stores the complete sequence of state-action pairs (i.e., events) leading to goal states (e.g., rewards) conserving their serial order, instead of storing world states and actions as isolated memory elements. A partial exception here is ERLAM, which does not treat events as being completely independent. However, the main difference between SEC and ERLAM is that whereas the latter builds a graph based on the state transitions of the stored experiences to bootstrap the learning of a parametric RL agent, SEC stores memories in a sequential goal-oriented manner, conserving the temporal structure of action, and using this memory buffer directly for action selection and control. 

Like other episodic control models \cite{Blundell2016, Pritzel2017}, SEC also follows a non-parametric approach as it uses an episodic tabular memory to store previously reinforced experiences and their predecessor states (e.g. behavioural sequences) to guide decision-making when encountering similar states. However, SEC deals with memory retrieval in a different way, through a combination of a perceptual similarity metric and a winner-take-all mechanism. Moreover, SEC computes the action-value function of a given state based on the combination of three factors: perceptual similarity between perceived and retrieved states, sequential bias between memory states, and discounted reward value. 

%Non-parametric algorithms such as MFEC, do not face the issue of bootstrapping and temporal credit assignment as in the gradient cases, nor the computationally costly tree search of model-based planning

%%And third, driving action selection through a memory-driven policy. [This is what any standard Episodic Control does]

\section{Methods}
\label{sec:methods}
\subsection{Sequential Episodic Control}
\label{sec:methods:sec}

\begin{figure}
  \centering
  \includegraphics[width=\textwidth]{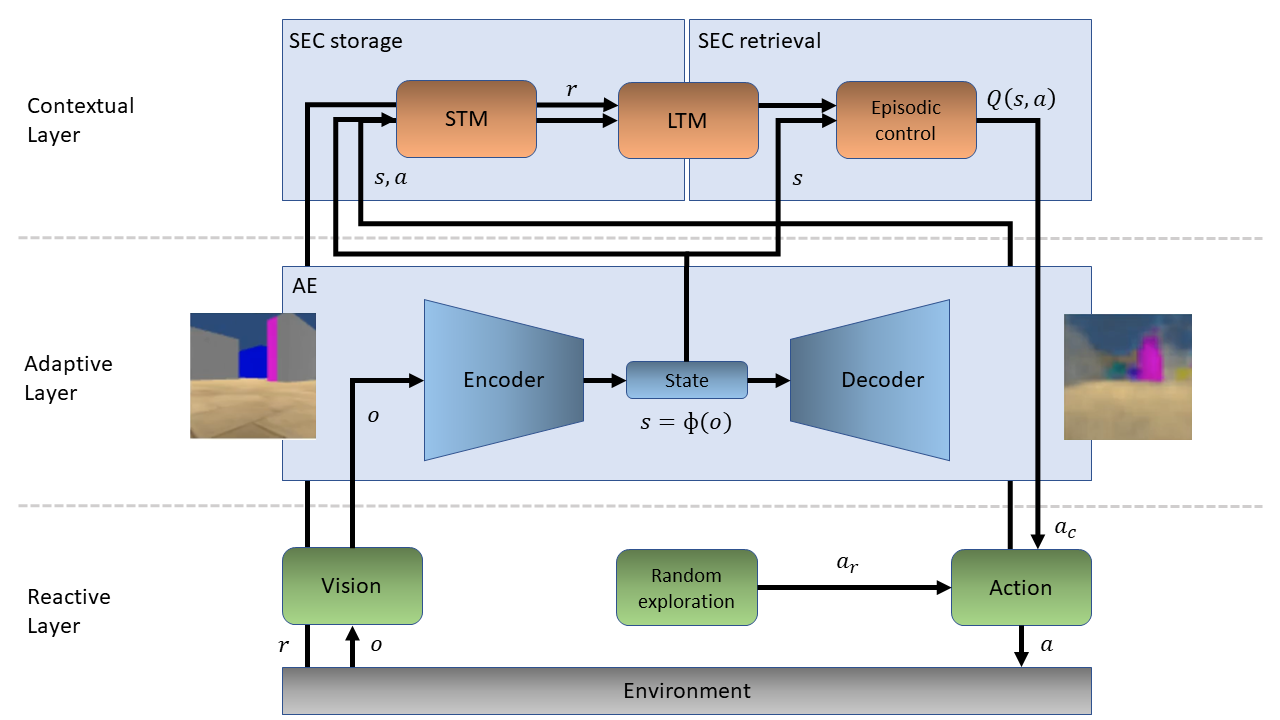}
  \caption{Sequential Episodic Control architecture. Following the Distributed Adaptive Control framework (\cite{Verschure2014} for review) SEC can be functionally divided into three layers: Reactive, Adaptive and Contextual. The reactive layer (green) implements a predefined random exploration algorithm. The adaptive layer (blue) acquires states of the world through a convolutional autoencoder, while the contextual layer (red) integrates a short- and long-term episodic memory buffer and an action-selection algorithm.}
  \label{sec_model}
\end{figure}

The SEC model is formulated in the context of the DAC theory of mind and brain \cite{Verschure2012}. DAC considers the brain as a multi-layer control system, comprising reactive predefined behaviors, adaptive state-space encoding, and contextual deliberation (i.e. the reactive, adaptive, and contextual layer, respectively; see Fig. \ref{sec_model}). The SEC model is fundamentally built on the consideration that the hippocampus incorporates systems for both the encoding of perceptual and action states and their integration into goal-oriented sequences, which correspond to the adaptive and contextual layers in the DAC framework \cite{Verschure2014}. Following this framework, SEC integrates the embedded states learned by its adaptive layer into the sequential memory system of the contextual layer. SEC's adaptive layer is composed of a convolutional autoencoder operating as an embedding function $ \phi $ that builds compressed feature representations $ s_t $ of the observations $o_t$ obtained from the environment $ s_t=\phi(o_t) $ (see appendix \ref{sec:appendix2:autoencoder} for more details). SEC's contextual layer includes a short-term memory buffer $STM$ and a long-term episodic memory $LTM$, combined with an action selection algorithm (see Algorithm \ref{alg:SEC}). Finally, SEC incorporates a random exploration algorithm in its reactive layer to drive the initial exploration of the state space and acquire its first memories. 

\begin{figure}
  \centering
  \includegraphics[width=\textwidth]{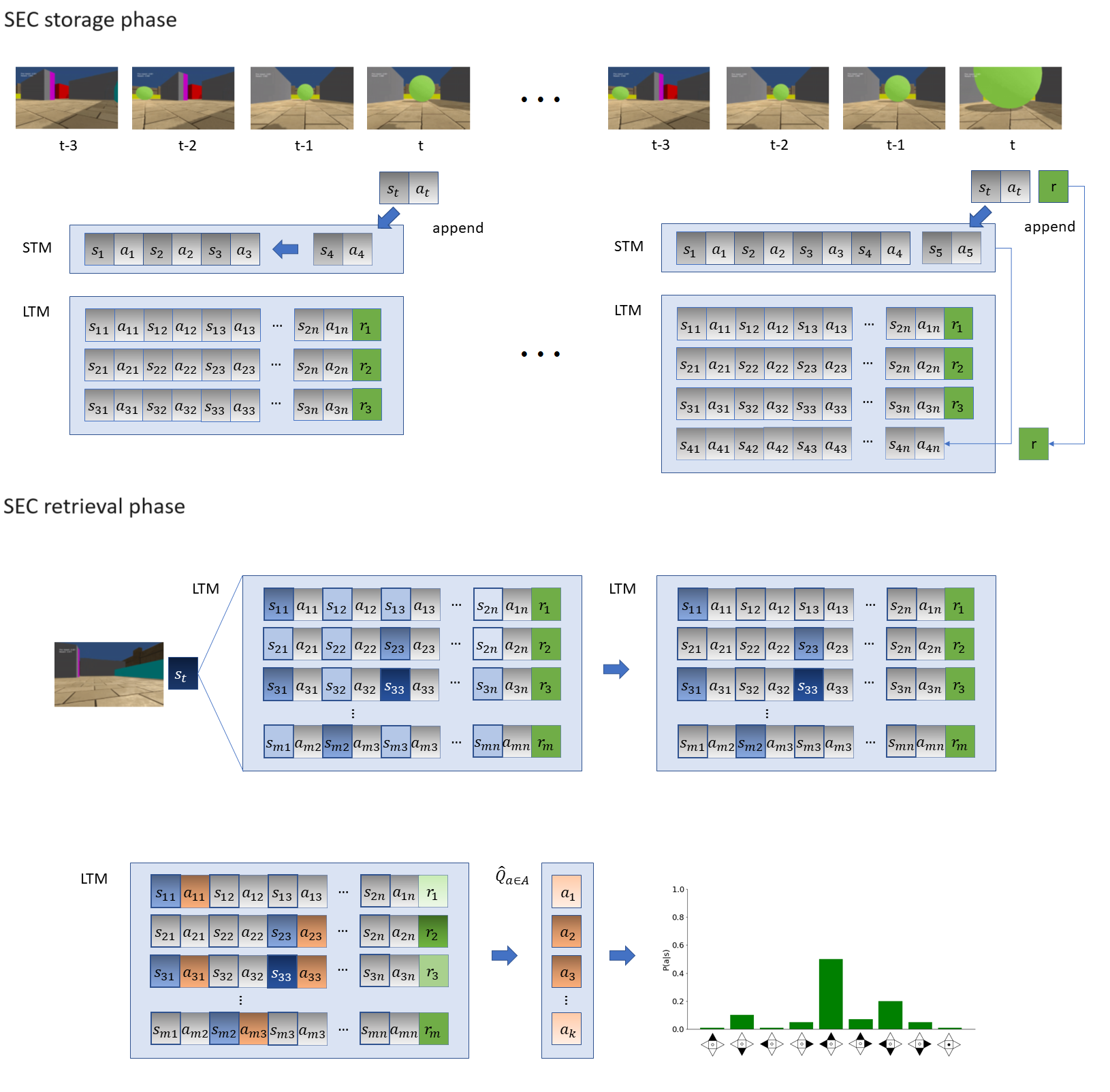}
  \caption{Sequential Episodic Control memory storage and retrieval phases. During the storage phase, state-action couplets are stored in the short-term memory (STM) on a first-in, first-out basis at every timestep (top-left). Upon encountering a reward, the content of the STM is transferred to the long-term memory buffer (LTM), along with the reward value (top-right). During the retrieval phase, first, following Eq. \ref{eq:eligibility}, the current observed state is compared with the stored states in the LTM and the most similar ones are retrieved (middle). After that, following Eq. \ref{eq:q_value}, the action-value function for that given observed state is computed by taking the actions attached to the retrieved states along with their discounted relative reward value (bottom).}
  \label{SEC_process}
\end{figure}

Episodic control algorithms have two main functions: memory storage and memory retrieval (see Fig. \ref{SEC_process} for a visual description). During the memory storage phase, the short-term memory buffer $STM$ transiently stores the most recent sequence of state-action pairs (i.e., events) encountered by the agent, and is updated following a first-in, first-out (FIFO) rule. Upon encountering a goal state (i.e., rewards), the sequence present in $STM$ is consolidated in $LTM$ along with its associated reward value ($r_t$). 

During the memory retrieval phase, memories are selected based on their perceptual and behavioral relevance to the current state of the agent. Concretely, for a given state ($s_t$), an eligibility score is computed for each state-action pair stored in $LTM$. The eligibility score $G_{i,j}$ of a state-action pair in memory is defined by its combined perceptual matching and sequential values, as expressed in Eq. \ref{eq:eligibility}. 

\begin{equation} \label{eq:eligibility}
\ G_{i,j}=(1-\ {d(s}_t,\ s_{i,j}))\ T \;, \;\; i,j \epsilon LTM \ 
\end{equation}

To calculate the perceptual matching, at every timestep, the perceived state ($s_t$) is compared, based on a similarity metric (i.e., distance $d(s_t,\ s_{i,j})$ based on the mean absolute error), to all the state representations stored in the $LTM$ (see Eq. \ref{eq:distances}). 

\begin{equation} \label{eq:distances}
\ {d(s}_t,\ s_{i,j})=\ \frac{1}{N}\ \sum_{k=1}^{N}\left|\ s_{t,k}-s_{i,j,k}\right| 
\end{equation}

The sequential matrix $T$ keeps track of the recent history of retrieved state-action memory pairs. At the beginning of each episode, $T$ is initialized as a unit matrix with dimensions equal to the $LTM$ ($m \times n$). At every timestep $t$, the sequential value of an element of a $LTM$ sequence is increased by $\alpha$ if the preceding element of such a sequence was previously selected at $t-1$ (see Eq. \ref{eq:sequential_bias}), indexed by the mask matrix $M$ (with $M_{i,j}=1$ for selected ones). Unselected memories (i.e., $M_{i,j}=0$) slowly return to their initial unit values at a constant rate of decay $\beta$. This mechanism implements a sequential inductive bias that favors the retrieval of complete memory sequences by orderly enhancing the selection of subsequent elements within a sequence over time.

\begin{equation}\label{eq:sequential_bias} 
\ T_{i,j}(t) = T_{i,j}(t-1) - \beta + \alpha \;\; \textrm{with} \;\; \alpha \neq 0 \;\; \textrm{if} \;\; M_{i,j-1}=1  \  
\end{equation}

This process reflects the sequential activation of hippocampal index neurons \cite{goode2020integrated}, during the phenomenon known as phase precession  \cite{skaggs1996phaseprecession}, and the consequential re-activation of cortical patterns representing the stored content of the events \cite{Lisman2009prediction}. The sequential bias is further amplified through behavioral feedback \cite{Verschure2003}, whereby the input sampling caused by SEC's deliberation process favors the storage of similar sequences of state-action pairs. 

To complete the retrieval phase, each memory has to go through a final selection process based on its eligibility score $G_{i,j}$. Only those state-action pairs that surpass both the absolute ($\theta_{abs}$) and proportional ($\theta_{prop}$) thresholds are retrieved to be used for action selection (see Eq \ref{eq:thresholds}, where $H(x)$ is the Heaviside function and $M$ is the resulting mask matrix with dimensions $m \times n$). This procedure enforces a soft winner-takes-all (WTA) mechanism akin to theta-gamma oscillatory dynamics in the hippocampus \cite{De_Almeida2009-emax}.

\begin{equation} \label{eq:thresholds} 
\ M_{i,j}=H\left(G_{i,j}-\theta_{abs}\right) H \left(\frac{G_{i,j}}{G_{max}}-\theta_{prop}\right)\ \ \ 
\end{equation}

In the action selection phase, the action-value function $\widehat{Q}_{a\epsilon A}$ is computed following Eq. \ref{eq:q_value} given the indexes of the selected state-action pairs $M$ from the memory retrieval phase. First, the value of each selected state-action pair $\left. Q_{s,a} \right|_{M}$ is computed by using its eligibility score $G_{i,j}$ and relative discounted reward. More concretely, the discounted reward is obtained by applying an exponential decay $e ^\frac{d_{i,j}}{\tau}$ to the relative reward of the corresponding memory sequence (i.e., $r_i/r_{max}$ where $r_{max}$ is the maximum reward across the selected pairs $M$). In turn, the decay is based on a time constant $\tau$ and the distance $d_{i,j}$ from the state-action pair to the end of the sequence. This mechanism implements a relative reward valuation between the selected memories \cite{tremblay1999relative,cromwell2005relative, soldati2017long}. Finally, the action value function $\widehat{Q}_{a\epsilon A}$ is the result of summing across the state-action values $\left. Q_{s,a} \right|_{M}$ for every action $a$ in the action space $A$.

\begin{equation} \label{eq:q_value} 
\ \widehat{Q}_{a\epsilon A} = \left. \sum_{i,j\epsilon M}{G_{i,j}\frac{r_i}{r_{max}}\ e ^{-d_{i,j}/\tau}}\ \right|_{a} \ \end{equation}

Then, the resulting action is selected by sampling the probability distribution generated by normalizing $\ \widehat{Q}_{a\epsilon A}$ so that $Q$ values sum up to 1. This method for computing the $Q$ values from relevant memory sequences favors the selection of actions that were taken in very similar states while prioritizing those that are closer to potential rewards. 

%The action selection algorithm selects an action based on the agent's recorded history of observed states and performed actions stored in $LTM$. This process is divided in two steps: memory retrieval and action selection. 

\begin{algorithm}
\caption{Sequential Episodic Control}\label{alg:SEC}

\begin{algorithmic}

\State $STM$: short-term memory 
\State $LTM$: long-term memory

\For{each episode} 

\State Initialize empty $STM$
\State $t = 1$

\While{$t < T$ and $r_t = 0$}
    \State Receive observation $o_t$ from environment
    \State Let $s_t=\phi(o_t )$
    \State Retrieve relevant memories for state $s_t$ via Eq.\ref{eq:eligibility}, Eq. \ref{eq:thresholds}
    \State Estimate return for each action $a$ via Eq.\ref{eq:q_value}
    \State Let $a_t \gets \pi(\widehat{Q}_{a\epsilon A}(s_t))$
    \State Take action $a_t$, receive reward $r_{t+1}$
    \State Append $(s_t, a)$ to $STM$
    \State $t \gets t + 1$
\EndWhile

\If{$r_t > 0 $}
    \State Append $(STM,r_t )$ in $LTM$
\EndIf

\EndFor

\end{algorithmic}
\end{algorithm}

\subsection{Experimental setup}
\label{sec:methods:exp}

This paper presents a sequence of four experiments aimed at evaluating the performance of the Sequential Episodic Control (SEC) model. Experiment 1 benchmarks the SEC model against established state-of-the-art algorithms across four challenging benchmarks from the Animal-AI testbed. Subsequent experiments focus on the Double T-Maze task to investigate the effects of memory capacity (Experiment 2), the contribution of individual components within the model's valuation function (Experiment 3), and the effect of forgetting mechanisms (Experiment 4). 

%Experiment 1 lays the groundwork by pitting the SEC model against a suite of state-of-the-art algorithms across four challenging benchmarks from the Animal-AI testbed. Building upon these findings, Experiments 2, 3, and 4 delve deeper into the SEC model's architecture, scrutinizing the effects of memory parameters and the implementation of forgetting mechanisms, all within the confines of the Double T-Maze task—a task selected for its ability to critically assess an agent's navigational acuity and memory retention. Each model was rigorously tested across 20 simulations, encompassing 5000 episodes each, to accumulate a substantial dataset of 5 million frames, ensuring a robust assessment of performance and stability. These experiments are designed not only to validate the SEC model's competencies but also to shed light on the potential enhancements that memory optimizations and strategic forgetting may confer.

%Experiment 1 benchmarks the SEC model across four tasks designed to test its cognitive and navigational capabilities. 

\textbf{Experiment 1} involves a comparative analysis against several state-of-the-art reinforcement learning and episodic control models, including the Deep-Q Network (DQN) \cite{Mnih2015}, Model-Free Episodic Control (MFEC) \cite{Blundell2016}, and Episodic Reinforcement Learning with Associative Memory (ERLAM) \cite{zhu_episodic_2020}. Moreover, in order to understand the specific role of sequentiality, we also added a control version of the model (NSEC) that does not incorporate the sequential bias $T$ (see Eq. \ref{eq:eligibility}). 

Regarding the models' hyper-parameters, a standard grid parameter search was performed to set the values of SEC. The same values were also used for NSEC. A detailed account of the hyper-parameters used by the SEC and NSEC models in this experiment can be found in \ref{sec:appendix1:hyperparameters}. For the MFEC algorithm, we reproduced the implementation described in \cite{Blundell2016} with $k = 50$ giving the best performance. We implement both versions of MFEC; MFEC-rp, which uses random projections as the embedding function, and MFEC-ae, which uses an autoencoder for the embedding function. The SEC, NSEC and MFEC-ae models use the same autoencoder architecture. As in previous approaches \cite{Blundell2016}, for each benchmark, we first train the autoencoder for 10.000 episodes (approximately 10 million frames) using random exploration, and then weights were frozen for the experimental phase (see \ref{sec:appendix2:autoencoder} for more details). Regarding the DQN, we keep the standard setting for network architecture and hyper-parameters as in \cite{Mnih2015}. Finally, for the ERLAM algorithm, we reproduce the algorithm following \cite{zhu_episodic_2020} using the same reported hyper-parameters (notably $\lambda = 0.3$). 

\begin{figure}[ht]
  \centering
  \includegraphics[width=\textwidth]{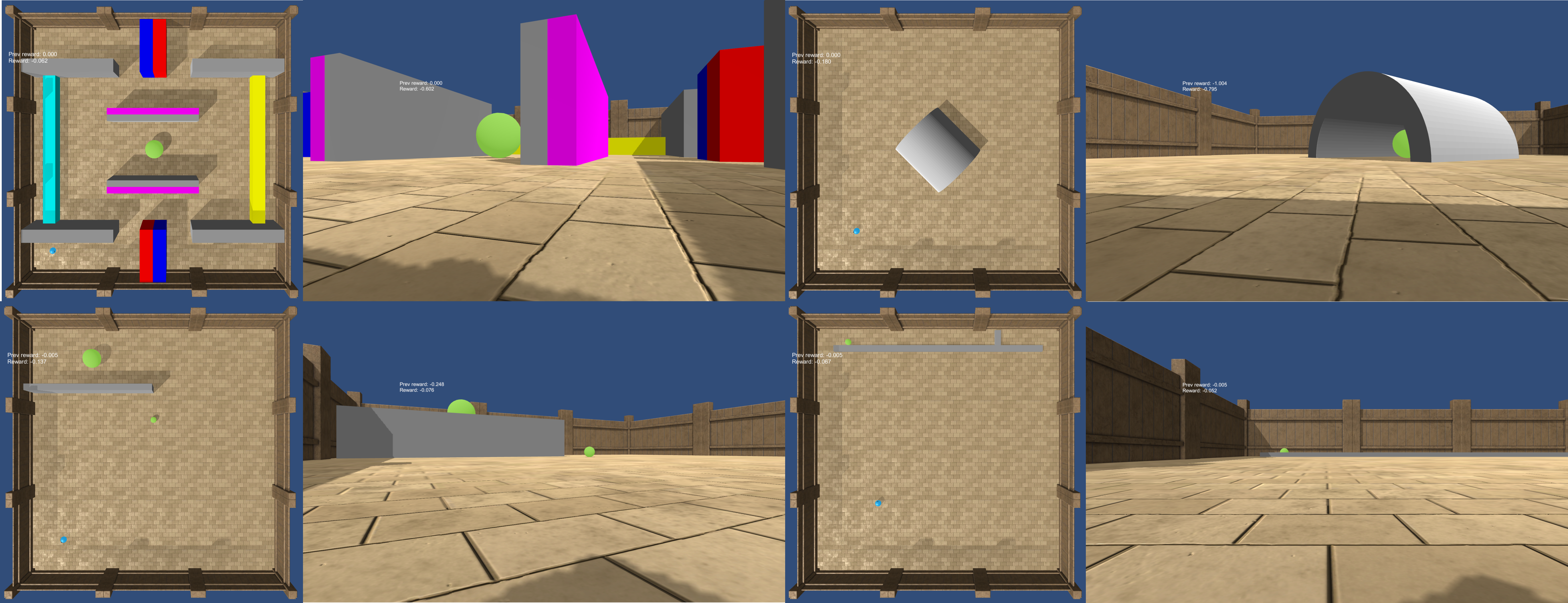}
  \caption{Illustration of the Animal-AI benchmarks, showcasing both overhead and agent perspectives. The displayed environments include Double T-Maze (top-left), Cylinder (top-right), Object Permanence (bottom-left), and Detour (bottom-right). For each benchmark, the left side of the panel provides a third-person, bird's eye view of the environment, while the right side offers the first-person perspective as seen by the agent navigating the scenario. }
  \label{SEC_exp_setup}
\end{figure}

In order to draw overall performance comparisons, we test all the models (SEC, NSEC, DQN, MFEC-ae, MFEC-rp, ERLAM) in four benchmarks of the Animal-AI testbed \cite{crosby2019animal, crosby2020animal}: The double t-maze, the detour task, the object permanence task, and the cylinder task (see Fig. \ref{SEC_exp_setup}).

\begin{itemize}
    \item The \emph{double T-maze} is a task with a sparse reward structure. This task requires the sequential encoding and retrieval of relevant visual cues to reach the reward location in a minimal amount of time. In each episode, the agent randomly starts at one of the corners, and it must reach the center of the maze to obtain a reward. The reward at the center is the only positive reward (+3) available in the environment. Due to the walls of the maze, the agent does not see the reward directly and needs to explore the maze first. 
    \item The \emph{object permanence} task involves food that moves out of sight that the agent needs to still attain. At the beginning of the episode, the agent can observe how a big reward (+3) falls into one of several holes of the maze until it is completely occluded. The agent needs to find the shortest path to reach the hidden reward while avoiding falling into the other holes. 
    \item The \emph{cylinder} task includes either opaque or transparent cylinders. In this task, the agent needs to get inside the cylinder to reach a medium-sized reward (+2). 
    \item The \emph{detour} task tests the ability to make a detour around an object to get food and assess the shortest path to the object. The wall is transparent but cannot be traversed, so the agent can perceive the reward from the other side. The small reward (+1) at the top corners is the only positive reward available in the environment. 
\end{itemize}

In all these environments, agents receive at each step the first-person visual scene from the environment with a resolution of 84 by 84 pixels. The action space is composed of a 2-d vector of integers that allows values from 0 to 2. The first value of the vector represents the translation axis, and the second value is the left-right axis. We apply a standard frameskip of 4 to reduce computational requirements. At each step, the reward value decreases by 0.001 to promote efficient trajectories. An episode finishes when the agent gets the reward or if 1000 frames are reached. 

%To test the generalization capacity of the SEC algorithm, we also test the models on the Atari Space Invaders benchmark  from the Arcade Learning Environment (ALE; \cite{bellemare2013arcade}), using the standard input pre-processing and enviromental dynamics presented in \cite{machado2018revisiting}: The raw images are resized to an 84 × 84 grayscale image, 4 consecutive frames are stacked into one state; frameskip is set to 4; episodes do not end with life; reward clipping is applied; and 30 no-op steps are performed at the beginning of each episode. For the evaluation, following the same guidelines, we report the full learning curve using the average total reward of each model over a sliding window of 100 episodes. 

\textbf{Experiment 2} was designed to investigate the influence of limited memory capacity on the performance of the Sequential Episodic Control (SEC) algorithm. We compared the standard SEC with its ablated version, NSEC, under various long-term memory (LTM) capacities in the Double T-Maze task. The memory capacities tested were fixed at 125, 250, 500, and 1000 sequences, meaning that once these capacities were reached, no additional memories could be recorded. This setup allowed us to examine the efficiency and effectiveness of memory use within these algorithms. 

In \textbf{Experiment 3}, we investigated the individual contribution of various mechanisms within SEC's valuation function (see Eq \ref{eq:q_value}) to its overall performance. This was accomplished by conducting ablation studies within the context of the Double T-Maze environment. We carried out two sets of ablation studies:

\begin{itemize}
    \item Single Mechanism Ablation: In the first set, we inactivated one mechanism at a time from the SEC's decision process, which includes the eligibility score $G_i$, the distance to the goal (Dist), and the relative reward (RR). This resulted in three reduced versions of SEC: SEC-noGi, SEC-noDist, and SEC-noRR, each lacking one of these respective mechanisms.
    \item Double Mechanism Ablation: In the second set, we inactivated two mechanisms simultaneously, allowing us to observe the performance when only one mechanism was active. This produced three variants: SEC-soloGi, SEC-soloDist, and SEC-soloRR, each utilizing a single mechanism from the action selection equation.
\end{itemize}

In \textbf{Experiment 4}, we extended our investigation of the Sequential Episodic Control (SEC) models to evaluate the effects of memory constraints and forgetting mechanisms on performance. Specifically, we explored how these factors influence the models' behavior in the Double T-Maze task, an environment that requires sophisticated memory management for optimal navigation and decision-making.

We introduced two types of forgetting mechanisms into the Long-Term Memory (LTM) of the models:
\begin{itemize}
    \item First-In, First-Out Forgetting (fifo): We applied a FIFO rule to the LTM, akin to the update mechanism of SEC's short-term memory (STM).
    \item Prioritized Forgetting (rwd): In this new condition, less rewarding memories are more likely to be forgotten, prioritizing the retention of high-reward experiences.
\end{itemize}

The models tested under these conditions included the original SEC and its non-sequential counterpart, NSEC, as well as their respective forgetting variants: SEC-fifo, NSEC-fifo, SEC-rwd, and NSEC-rwd. 
%We conducted 20 simulations per model, each running for 5 million frames.

Each reported experiment involved 20 simulations per model, with each simulation running for 5000 episodes (5 million frames), to ensure statistical reliability. These controlled experiments are designed to dissect the operational parameters of the SEC model, providing insights into its functionality under different conditions and contributing to the broader understanding of episodic control in artificial intelligence. 

\section{Results}
\label{sec:results}

\subsection*{Experiment 1. Sequentiality improves performance of episodic control}
\label{sec:results:exp1}

\begin{figure}[!ht]
  \centering
  \includegraphics[width=\textwidth]{fig4_all_benchmarks.png}
  \caption{Comparative performance of Sequential Episodic Control (SEC) against several benchmark algorithms, namely Deep-Q Network (DQN), Model-Free Episodic Control (MFEC), Episodic Reinforcement Learning with Associative Memory (ERLAM), and non-sequential ablated version of SEC (NSEC). The presented results encompass four distinct Animal-AI benchmarks; the Double T-Maze (top-left), Cylinder (top-right), Object Permanence (bottom-left), and Detour (bottom-right) tasks. For clarity and statistical robustness, average performance metrics were calculated using a sliding window encompassing 20 episodes (20.000 frames). The error bars denote the standard error (SE) to provide a measure of the variability in the dataset.}
  \label{SEC_main_results}
\end{figure}

Experiment 1 evaluated the Sequential Episodic Control (SEC) model across four benchmarks: Double T-Maze, Object Permanence, Cylinder, and Detour tasks within the Animal-AI environment. The performance of SEC was compared against several state-of-the-art models including Non-Sequential Episodic Control (NSEC), Model-Free Episodic Control with an autoencoder (MFEC-ae), Model-Free Episodic Control with random projections (MFEC-rp), Deep Q-Network (DQN), and Episodic Reinforcement Learning with Associative Memory (ERLAM).

In the Double T-Maze task, the SEC model demonstrated superior performance, achieving higher average rewards faster and sustaining those rewards over time compared to the other models. This benchmark emphasizes the model's proficiency in tasks requiring the encoding and retrieval of sequential information to navigate toward a goal.
%In the Double T-Maze task, the SEC algorithm shows a strong and stable learning curve, achieving a higher average reward and stabilizing faster than the other models. This indicates that SEC is particularly adept at handling complex pathfinding tasks that require sequential decision-making. ERLAM and DQN achieved the second-best results on this task, with MFEC-ae and NSEC following. 

For the Object Permanence task, which tests the model's ability to remember and act upon the location of unseen objects, SEC again outperformed the competing algorithms, showing a rapid increase in average rewards. This indicates the model's effectiveness in scenarios where indirect cues must guide decision-making. The results of DQN and ERLAM seem to indicate that they have fallen into the local minima of capturing the small visible reward (+1), instead of pursuing the hidden but greater reward (+3), which was more difficult to find. 

%For the Object Permanence task, the SEC algorithm also demonstrates superior performance, with a more rapid increase in average reward than the other tested models. The results of DQN and ERLAM seem to indicate that they have fallen into the local minima of capturing the small visible reward (+1), instead of pursuing the hidden but greater reward (+3), which was more difficult to find. The two versions of MFEC and NSEC show better performance than DQN and ERLAM, indicating their capacity to reach the optimal solution to the task, although they do not reach the performance levels of SEC 

The Cylinder task, requiring the discernment between opaque and transparent obstacles, saw a similar trend, with the SEC model matching or outperforming other models, albeit with a closer margin. The results suggest that while SEC is adept at tasks requiring visual discrimination, the advantage is less pronounced in this context.

%The Cylinder task results show that SEC achieves better results than comparable rewards but with noticeably less variability in performance than the other top-performing models NSCE, MFEC-rp, and ERLAM. 

In the Detour task, designed to assess the ability to plan and execute a path around an obstruction, SEC outshone all other models significantly, as evidenced by the steep and consistent rise in average reward. The SEC model's performance in this task underscores its capacity for handling complex spatial navigation challenges.

%Finally, in the Detour task, SEC exhibits a distinct advantage, with the reward curve climbing steeply and plateauing at a higher level compared to other models, which are not able to succeed in this task. This task seems to underscore the benefits of the sequential approach that SEC embodies, as it shows that in a hard exploration task where there are few instances in which the models can reach the reward by chance, models like SEC can provide a fast way to quickly latch onto successful strategies. 

Across all tasks, the NSEC model performed notably worse than its sequential counterpart, emphasizing the critical role of sequentiality in the SEC model's success. The MFEC variants, while competitive, did not reach the performance peaks of SEC. DQN and ERLAM trailed behind, particularly in tasks that demanded more sophisticated episodic memory capabilities.

%The comparison between SEC and its non-sequential counterpart, NSEC, across these benchmarks provides clear empirical evidence that the sequential chaining bias integrated into SEC is a potent mechanism for enhancing its performance. The findings suggest that this bias allows SEC to better generalize its experience through the chaining of episodes, leading to more efficient learning and problem-solving. The overall analysis conveys that the sequential inductive bias in SEC is a critical component for its success, enabling it to achieve higher rewards more efficiently than its contemporaries and variants without this feature.

%A particularly interesting aspect of these results is the marked improvement in SEC's performance when compared to its non-sequential variant, NSEC. This variant lacks the sequential inductive bias inherent in SEC. The comparative analysis between SEC and NSEC highlights that the notable performance enhancement in SEC is directly attributable to its sequential inductive bias, underscoring the efficacy of this feature in the model's overall success.

The results from Experiment 1 underscore the SEC model's robustness and versatility across a variety of tasks that challenge different cognitive skills. The consistent outperformance of SEC over NSEC and other models across all benchmarks highlights the efficacy of sequential episodic control mechanisms in complex navigational and cognitive tasks.

%Taken together, these findings reveal that the Sequential Episodic Control (SEC) model consistently surpasses the other tested state-of-the-art models in all four benchmarks (see Fig. \ref{SEC_main_results}). Notably, SEC not only achieves higher reward levels but also reaches these performance peaks more rapidly, showcasing its superior sample efficiency.

%The results show that the SEC model clearly outperforms the other state-of-the-art models in the four benchmarks (see Fig. \ref{SEC_main_results}). SEC reaches not only higher levels of reward but also gets faster to such level of performance, thus making it more sample-efficient. Interestingly, we can also observe that there is a clear improvement in the performance of the SEC model in comparison to SEC's ablated version, NSEC, which does not possess the sequential inductive bias. These results demonstrate that the significant increase in the performance of SEC against NSEC is due to its sequential inductive bias.

%In the Space Invaders benchmark, SEC achieves superior performance than DQN and closely matches ERLAM. However, in this case, MFEC outcompetes the other models. These results show under two qualitatively different benchmarks, SEC is able to consistently achieve state-of-the-art performance, showing less variability between the tested environments than other episodic control algorithms like MFEC. 

\subsection*{Experiment 2. Sequential bias enhances performance and memory efficiency in SEC under memory constraints}
\label{sec:results:exp2}

\begin{figure}[!ht]
  \centering
  \includegraphics[width=\textwidth]{fig5_tmaze_memories.png}
  \caption{Effect of memory constraints across episodes between SEC and NSEC models in terms of reward accumulation and entropy in the Double T-Maze benchmark. Top panels: Mean reward per episode accumulated by SEC (left) and Non-SEC models (right). Bottom panels: Mean entropy on the episodic policy, computed as the average of the entropies of the probability distributions derived from $\ \widehat{Q}_{s,a} ^{LTM}$ at every timestep of the episode. Vertical bars represent the average episode around which the memory was filled. Average values were computed using a sliding window of 20 episodes. Error bars represent SE.}
  \label{SEC_memory_results}
\end{figure}

To analyze the effect that limited memory capacity imposes on the Sequential Episodic Control algorithm, we tested the standard (SEC) and ablated (NSEC) versions of the model with varying long-term memory capacities (i.e., a fixed LTM memory of 125, 250, 500 and 1000 sequences) in the Double T-maze task. Therefore, once the memory was filled, no further memories could be stored. As in Experiment 1, we performed 20 simulations of 5000 episodes (approximately 5 million frames) per model and memory condition.

The results of Experiment 2 show that SEC maximizes reward acquisition upon reaching one thousand episodes for all tested memory conditions (see Fig.\ref{SEC_memory_results}). In terms of performance, the SEC model obtains a clear advantage with respect to its non-sequential version, NSEC, across all memory conditions. As before, SEC reaches not only higher levels of accumulated reward but also reaches asymptotic performance more quickly. The difference in convergence rates to fill the sequential memory illustrates the bootstrapping effect of behavioral feedback. This difference increases between the two versions of the model when the memory capacity limit is increased. In all cases, the performance plateaus shortly after reaching the memory capacity (as indicated by the vertical bars in Fig.\ref{SEC_memory_results}). Importantly, the decrease in mean entropy shows that SEC stabilizes its policy, whereas NSEC does not, maintaining a high policy entropy in all conditions despite increasing reward acquisition. Thus, the sequential bias component of SEC allows the agent to achieve a higher level of accumulated rewards and also provides a behavior stabilization mechanism.

\begin{figure}[!ht]
  \centering
  \includegraphics[width=0.5\textwidth]{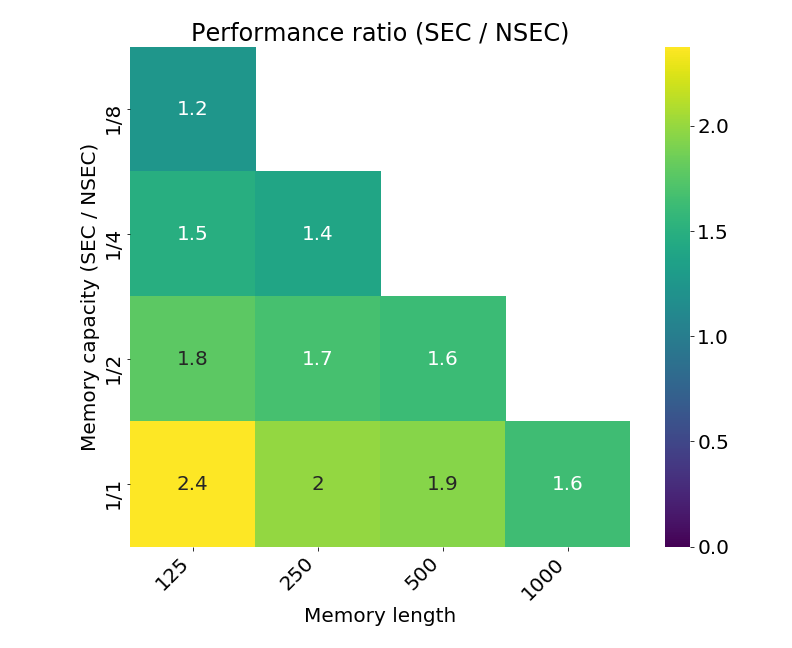}
  \caption{Performance increase of SEC over NSEC across different memory-limit conditions in the Double T-maze benchmark. Units reported are the total mean performance of SEC over NSEC. Each column shows SEC's performance with a limited memory capacity of 125, 250, 500, and 1000 sequences respectively. Each row shows the memory ratio between SEC and NSEC, ranging from 1/1 (equal memory limit) to 1/8 (SEC memory limit is 8 times smaller than NSEC).}
  \label{SEC_memory_matrix}
\end{figure}

The analysis of the policy entropy of the SEC model shows similar patterns to the reward accumulation dynamics, indicating that agents using the SEC model quickly converge to a stable policy. The entropy of the NSEC model, on the other hand, is much higher than SEC in all conditions. It is also important to note that the degree and speed of convergence are significantly affected by memory capacity. In both episodic control models, there is a consistent reduction in entropy when the memory capacity is increased. This effect indicates an improvement in the stabilization of the policy that also is reflected in the increased reward acquisition.

Finally, we systematically compare the performance of each memory condition of NSEC against SEC to have a detailed estimate of how much performance increase is obtained using the sequential bias. The results shown in Fig.\ref{SEC_memory_matrix} present a table comprising this analysis. The data shows that the SEC model can reach a similar level of performance to its non-sequential version, NSEC, with an order of magnitude lower memory capacity. In other words, the SEC model with a fixed memory of 125 units achieves 1.2 times the performance of NSEC with 1000 memory units. Moreover, when both models are evaluated with equal memory limitations, the performance difference between SEC and NSEC increases as the memory capacity is reduced: SEC obtains 1.5 times the performance of NSEC with a limit of 1000 memories, and it scales up to 2.5 times with 125 memory units. These results show the key advantage of sequential episodic control algorithms because the sample efficiency of SEC also translates into memory efficiency and, therefore, reduces computational costs. This demonstrates the favorable scaling properties of SEC and similar solutions.

\subsection*{Experiment 3. No strong impact of individual valuation components on Sequential Episodic Control performance}
\label{sec:results:exp3}

\begin{figure}[!ht]
  \centering
  \includegraphics[width=\textwidth]{fig7_tmaze_ablations.png}
  \caption{Ablation studies of the Sequential Episodic Control (SEC) algorithm in the Double T-Maze task. The left panel presents the single mechanism ablations with average reward (top) and entropy (bottom). The right panel shows the double mechanism ablations under the same metrics. In the single ablations, 'SEC-noDist' lacks the distance to the goal component, 'SEC-noRR' lacks the relative reward component, and 'SEC-noGi' lacks the eligibility score component. In the double ablations, 'SEC-soloDist', 'SEC-soloRR', and 'SEC-soloGi' operate with only the distance to the goal, relative reward, and eligibility score components, respectively. The full SEC model is included as a benchmark in both panels. These graphs demonstrate the comparative impact of individual and combined components of the SEC algorithm on its performance and decision-making uncertainty. Vertical bars represent the average episode around which the memory was filled. Average values were computed using a sliding window of 20 episodes. Error bars represent SE.}
  \label{SEC_ablation_results}
\end{figure}

In this study, we perform several ablations on SEC's valuation function (see Eq \ref{eq:q_value}) in order to analyze the differential effect of its three components: the eligibility score  $G_i$, the distance to the goal, and the relative reward. The results of the ablation studies are depicted in two graphs in Fig. \ref{SEC_ablation_results}, one for each set of studies, the Single Mechanism Ablation and the Double Mechanism Ablation. 

In the Single Mechanism Ablation study, omitting the eligibility score $G_i$ (SEC-noGi) resulted in a slight decrease in average reward compared to the full SEC model, suggesting that while $G_i$ contributes to performance, its absence doesn't drastically impair the model. When the distance to the goal was not considered (SEC-noDist), we observed a more pronounced drop in the average reward, indicating that spatial considerations play a more significant role in SEC's success. Finally, the removal of the relative reward component (SEC-noRR) had an intermediate impact, more than SEC-noGi but less than SEC-noDist, underscoring its importance but also the model's resilience to its absence. The entropy plots correlate with these findings, showing that the models with a single mechanism removed generally maintained similar uncertainty levels in their action selection.

The Double Mechanism Ablation results indicate that the SEC model retains competitive performance even when reduced to a single operational mechanism, be it the eligibility score  $G_i$ (SEC-soloGi), distance to the goal (SEC-soloDist), or relative reward (SEC-soloRR). Throughout the ablation studies conducted in this experiment, it's noteworthy that although the modified versions did not achieve the same high level of results as the complete SEC model, they consistently outperformed the state-of-the-art models evaluated in Experiment 1 within the Double T-Maze task. 

In contrast with these results, NSEC (the variant of SEC where the sequential bias is removed) obtained significantly inferior results compared to all the other ablated versions and the complete SEC model. This marked difference highlights the pivotal role of sequentiality within SEC. The sequential bias is evidently a critical factor that contributes to the SEC's optimal performance in complex tasks such as the Double T-Maze, and its absence is detrimental to the model's success. In essence, while the SEC can function above state-of-the-art standards without certain components, it is the integration of sequentiality that propels it to achieve the best results.

\subsection*{Experiment 4. Forgetting enhances performance and policy stability}
\label{sec:results:exp4}

\begin{figure}[!ht]
  \centering
  \includegraphics[width=\textwidth]{fig8_tmaze_forgetting.png}
  \caption{Forgetting enhances episodic control performance and policy stability in the Double T-Maze benchmark. Model comparison of episodic control models with forgetting (SEC-FIFO: dark green, NSEC-FIFO: orange), without forgetting (SEC: green, NSEC: red) against the MFEC (blue) benchmark. Left: Average performance per episode. Right: Entropy over the policy for SEC and NSEC models. Lower entropy values imply greater policy stability. Average values were computed using a sliding window of 20 episodes. Error bars represent SE.}
  \label{SEC_forgetting_results}
\end{figure}

The results from Experiment 4, depicted in Fig.\ref{SEC_forgetting_results}, indicate that the integration of forgetting mechanisms has a nuanced impact on the performance of episodic control models. 

%The SEC models described in Experiments 2 and 3 were tested with a fixed LTM buffer to assess the effect of memory limitations on the performance of the models. In Experiment 4, we study how adding a forgetting mechanism affects the models' performance in the Double T-Maze task, and how it interacts with the sequential chaining. Concretely, we study two forgetting conditions: FIFO, where we add a first-in, first-out rule to the long-term memory, similar to the one used to update SEC's short-term memory. For this experiment, 20 simulations of 5000 episodes (5 million frames) were performed for each of the models: SEC, NSEC, SEC-fifo, NSEC-fifo, SEC-rwd and NSEC-rwd.

For the SEC model, both forgetting mechanisms (SEC-fifo and SEC-rwd) demonstrate slightly enhanced performance over the original SEC, with the reward-based forgetting (SEC-rwd) variant showing a particularly notable improvement. This improvement is reflected in the average reward curves, where SEC-rwd achieves a higher average reward compared to both the original SEC and the SEC-fifo variant.

In the case of NSEC, the addition of forgetting (NSEC-fifo and NSEC-rwd) also results in improved performance, with the reward-based forgetting variant (NSEC-rwd) again standing out with a higher average reward than the NSEC-fifo.

%The results of Experiment 4 show that forgetting slightly increases the overall performance of both episodic control models (see Fig.\ref{SEC_forgetting_results}). The effects of forgetting can also be observed in the reduced entropy observed on the policy distribution of SEC-fifo and NSEC-fifo when compared to their fixed-memory counterparts. These results can be accounted for by how forgetting affects episodic control models, whose performance is entirely determined by the type of memories they are able to store. As shown in Fig.\ref{SEC_forgetting_results}, by the time the long-term memory of SEC-fifo and NSEC-fifo has been filled, the performance of the episodic controllers has already improved with respect to the early stages of the experiment. The ability to forget memories of earlier and less efficient behavioral sequences gives room for the storage of recent successful experiences. These newly acquired sequences, in turn, will tend to render higher returns since they are already capitalizing on the acquired experiences stored in memory. 

The entropy plots reveal that the addition of forgetting mechanisms leads to reduced entropy in the policy distribution, suggesting more deterministic and possibly more efficient behavior. This effect is more pronounced in the reward-based forgetting variants of both SEC and NSEC, indicating that prioritizing the forgetting of lower-reward experiences leads to a more focused and effective policy.

Overall, the performance enhancement from forgetting is evident but not as substantial as the improvement gained from sequential information processing, as seen when comparing the full SEC model to the NSEC variants with forgetting. This comparison underscores the critical role of sequentiality in the model's success. Nonetheless, the combined effects of forgetting and sequentiality contribute to the highest overall performance and the lowest entropy, as demonstrated by the SEC-rwd model. This suggests that the benefits of forgetting are additive when paired with the sequential chaining capability, resulting in an even more powerful episodic control model.

%The introduction of forgetting also allows us to assess the distinct contributions of sequentiality and forgetting in episodic control. Whereas the introduction of forgetting does improve the overall performance of the episodic controllers (see NSEC-fifo vs NSEC or SEC-fifo vs SEC results), it is not comparable to the performance and stabilization obtained by the use of sequential information in decision-making (see SEC vs NSEC-fifo). These results also show that the benefits of forgetting and sequentiality are additive, as the model integrating both features achieves the highest overall performance while maintaining the lowest entropy over its policy (SEC-fifo).

In summary, the introduction of forgetting mechanisms, particularly prioritized forgetting, improves the performance of episodic control models in the Double T-Maze task. However, it is the integration of sequentiality that most significantly enhances the model's capabilities, with the combination of both features yielding the best results.

%The introduction of forgetting also allows us to test whether the non-sequential model with the addition of a FIFO forgetting mechanism is able to reach SEC-level performance.
%We observe that the NSEC-FIFO algorithm does not yield the same results as applying a sequential bias, like in SEC, with a fixed LTM buffer . 
%However, in the case of SEC this improvement is not that significant. This might be due to the fact the SEC reaches almost optimal performance on this task with its current fixed LTM buffer limit.

\section{Discussion}
\label{sec:discussion}

Episodic control models seek to overcome the sample-inefficiency problem in Reinforcement Learning by implementing a form of instance-based learning that capitalizes on the storage of previously successful experiences to quickly learn optimal strategies. Although they are generally inspired by the mammalian hippocampus, those models treat stored experiences as isolated units, without taking into account the sequential nature in which those events unfolded in real time. In this paper, we show that the omission of this feature has implications both in terms of sample efficiency and memory efficiency for Episodic Reinforcement Learning. We present a novel episodic control algorithm, Sequential Episodic Control (SEC), that addresses both issues by incorporating a more complete picture of the hippocampal function. Crucially, SEC stores complete behavioral sequences in its long-term memory and takes into account the sequential nature of its stored experiences for action selection. We show how SEC can reach sample-efficient performance compared to episodic controllers that store state-action pairs as independent elements in memory, such as Model-free Episodic Control. To address memory efficiency, we systematically study the effect of memory limitations and forgetting in episodic control. We demonstrate that constraining the capacity of the memory buffer significantly affects the performance and stability of episodic control algorithms. The results show how the SEC model outperforms an episodic controller that does not take into account the sequential structure of the stored memories, and that this difference in performance is increased when greater memory limitations are imposed. Our results also indicate that forgetting generally improves performance and policy stabilization on episodic control, but sequentiality plays a major role in comparison. Taken together, this work seeks to contribute to the available proposed solutions to the sample-inefficiency problem in reinforcement learning by taking inspiration from empirical and theoretical research in the cognitive sciences. This work also extends the possibilities of episodic control algorithms by showing how leveraging the sequential nature of the stored experiences improves sample and memory efficiency during learning. 

%% MAIN DIFFERENCE BETWEEN MFEC AND SEC
In contrast with standard episodic control models, which treat their memory units as disconnected events, SEC emphasizes the learning of complete goal-oriented behavioral sequences. The incorporation of this hippocampus-based architectural bias bootstraps the formation of a behavioral feedback loop \cite{Verschure2003}, where the agent will tend to retrieve and orderly follow previously successful behavioral sequences, thus resulting in the generation of similar behavioral outcomes, which in turn will lead to the acquisition similar sequential memories. This memory-driven behavioral feedback loop allows SEC to rapidly transition from an initial exploration phase to an exploitation phase in which the behavior of the agent stabilizes over time while maximizing reward acquisition.

%%% MODEL-BASED AND MODEL-FREE EPISODIC CONTROL
%Within episodic control models, a theoretical distinction has been made between episodic model-based  and episodic model-free algorithms that resonate with the classical distinction in reinforcement learning \cite{vikbladh2017episodic}. In this view, both ERLAM and SEC will model-based episodic control system since both of the retain the sequential structure of the memory, and therefore keep certain order. In contrast, pure model-free episodic controllers would not make use of the temporal structure of the stored memories, like NSEC, or directly store them as discretised elements, as in MFEC \cite{Blundell2016}.

Within the framework of episodic control models, a theoretical distinction aligns with the traditional dichotomy in reinforcement learning between model-based and model-free algorithms \cite{vikbladh2017episodic}. In this case, the distinction is made based on the structure of the stored memories. According to this perspective, both ERLAM and SEC can be classified as 'episodic model-based' control systems because they preserve the sequential structure of memory, thereby maintaining a certain order of events. In contrast, pure 'model-free episodic' controllers, such as NSEC, do not utilize the temporal structure of stored memories, or they discretize and store events as isolated instances, as in MFEC \cite{Blundell2016}.

%%% MAIN DIFFERENCE BETWEEN ERLAM AND SEC
It is important to note that although SEC and ERLAM are both 'episodic model-based' systems, as they keep structure in the sequences of events they store, there are two key differences among them. The main difference between SEC and ERLAM is precisely that SEC makes use of the sequences for \textit{direct control}, whereas ERLAM makes use of the value estimates generated by its associative memory system as a \textit{target signal }for a DQN. Moreover, in ERLAM, there is an active associative process in which a model is built based on the stored memories to create a graph network over the visited states, effectively 
creating a model of the world, similar to model-based reinforcement learning methods \cite{zhu_episodic_2020}. An interesting avenue for future research would be to develop an episodic model-based algorithm that uses ERLAM's associative memory for direct control. 

While SEC and ERLAM are both categorized as 'episodic model-based' systems due to their retention of structured event sequences, they diverge significantly in application. The primary distinction is that SEC employs these sequences for direct control, whereas ERLAM utilizes the value estimates from its associative memory as a target signal for a DQN. Additionally, ERLAM actively engages in an associative process where a model is constructed from stored memories, creating a graph network over the visited states and effectively forming a world model akin to model-based reinforcement learning methods \cite{zhu_episodic_2020}. A promising avenue for future research involves a comparative analysis of SEC with an episodic model-based algorithm that incorporates ERLAM's associative memory for direct action selection. Such a study could elucidate the distinct contributions and roles of episodic memory and associative mechanisms within the decision-making process, providing deeper insight into the interplay between these cognitive systems in guiding behavior.

%For this reason SEC, building on the DAC framework, proposes that model-free and model-based systems constitute control layers of a single control architecture where the former also provides the content for the latter. In the case of SEC, this is the autoencoder of the adaptive layer driving the perceptual content of the sequential representations of the contextual one.

%% USING EPISODIC CONTROL FOR BOOTSTRAPPING LEARNING OF MF-MB
Conversely, although in this paper we have only shown the capacity of the SEC model for control, it could also be implemented like ERLAM's associative memory, that is, in combination with other learning algorithms to bootstrap the learning phase from a task-relevant set of successful samples, as shown in previous work \cite{Lin2018, Lee2018}. Moreover, such an application of the combination of control and batch learning does not need to be mutually incompatible. Indeed, it could be possible to use episodic control models to drive an agent's behavior during the initial stages of learning, by rapidly latching onto successful experiences, while a slow learning algorithm such as a DQN could use the growing set of successful sequences obtained by the episodic controller to speed up the acquisition of an optimal policy through offline batch learning and replay \cite{Mattar2018, caze2018hippocampal}.

%%% NEW: WHY SEC IS INSTANCE-BASED LEARNING AND NOT MODEL-BASED RL
Episodic control algorithms, by their inherent design, resonate deeply with the instance-based learning (IBL) theory from cognitive science \cite{gonzalez2003instance, gonzalez2011instance}. The IBL theory postulates that decisions are made based on the recall of specific past episodes or instances, rather than by aggregating across them \cite{gonzalez2003instance}. Similarly, episodic control models prioritize recent experiences, leveraging specific memories to make informed decisions \cite{Lengyel, Blundell2016}. These models can be seen as an emulation of the cognitive processes underlying episodic memory, where past instances are recalled to provide context and guide current decision-making \cite{Gershman2017, gonzalez2011instance}.

%Following this distinction between incremental learning rules such as MB adn MF reinforcement learning, "learning rules that extract statistical summaries from experience. Another possibility is that individual events are stored as episodic memories and later sampled to guide choice. Such episodic evaluation may confound standard tests for model use, since individual trajectories contain the same information as the map."

%Although it is true that by its nature, the sequential memories carry with them certain intrinsic structure of the world, and therefore such stored episodic trajectories might confound standard tests for model use, since in a way contains the same information as the map, that is only the case for a small fraction of such map from which that sample is stored. Regardless of this fact, episodic control model do not strictly build and explicit model of the world. It only capture instances, small fractions of it. In this sense, SEC is not fully a model-based episodic control algorithm, neither it is completely model-free like in the case of MFEC \cite{Blundell2016}. This instance-based memory system fall precisely in between those two cagetories. SEC directly use this past episodes of experience directly for decision-making.

%% DIFFERENCE WITH RL + MEMORY AND FORGETTING
In contrast with standard RL methods, the main driver of policy learning in episodic control is the constant acquisition and retrieval of new sequential memories. In other words, the behavioral policy of an episodic controller is implicitly updated by the memories it forms during its interaction with the environment. Precisely due to this factor, the performance of episodic control algorithms is sensitive to memory constraints (i.e. limits on its memory capacity), and therefore can also benefit from the implementation of adequate forgetting mechanisms, as shown in this work, and studied in \cite{Yalnizyan2021forgetting}. Therefore, in the absence of forgetting, when an episodic controller fills its memory capacity, its learning and performance will tend to stagnate since it will no longer be able to store new memories. 

Choosing when and what to forget are fundamental challenges for the generalization, robustness, and long-term performance of episodic control algorithms. Of course, problems regarding memory limitations and efficiency might not be relevant for models using unbounded memory buffers. Nonetheless, such issues become fundamental in the development of embodied artificial agents. In robotics and embodied AI, the question of autonomy is central, hence, the need for fast learning is intertwined with the optimization of energetic and computational demands, making solutions like sequential episodic control a promising avenue for progress.

The capacity of episodic control algorithms like SEC to quickly exploit past successful strategies can also be very useful in multi-agent environments to help in dealing with the continually changing policies of learning agents \cite{Papoudakis2019dealing, freire2020modeling}. Moreover, this type of sequential memory can also help in the formation of better internal models of other agents \cite{albrecht2018autonomous, freire2023modelingtom, freire2018limits, freire2018modelingtom}, by building a sequential memory of past social interactions supporting the virtualization of the "other" which is necessary for effective multi-agent and social interaction \cite{verschure2016synthetic, freire2020machine, freire2023high}.

%[I am missing a paragraph on potential shortcomings of the study and relation to other models including DAC.] - Done
%% LIMITATIONS OF EPISODIC CONTROL AND SEC
A notable challenge faced by this work, as well as by episodic control models more broadly, lies in their so-far limited domain of application \cite{Blundell2016}. Episodic control excels in environments with deterministic state transitions and rewards, and in situations where an agent can benefit from memories of similar past experiences \cite{Blundell2016, Pritzel2017}. However, further work is required to assess how well this type of algorithm performs in more complex, non-stationary settings, and how well it deals with generalization. One potential solution for managing non-stationary environments in episodic control is the development of more sophisticated forgetting mechanisms that can selectively prune outdated memories of events or states that are no longer relevant due to sudden changes in the environment. The capacity of episodic controllers to generalize their acquired knowledge to different tasks and environments depends on their ability to adequately recruit their memories for use in similar perceptual states. In this work, we show that SEC can perform very well in high-dimensional state spaces, where it can utilize its memories to generalize previously successful behaviors to similar states. SEC addresses state generalization by building internal state representations through a convolutional autoencoder and selecting memories based on the perceptual similarity between observed and stored states, but other options exist \cite{Yalnizyan2021forgetting}. Future research into different methods for building such internal representations might play a key role in developing more efficient episodic control algorithms.

%% MULTIPLE LEARNING MECHANISMS 
In nature, complex organisms make use of different learning, memory, and decision systems adaptively \cite{Verschure2012}. A classical distinction in cognitive science is between deliberate and habitual behavior. On the one hand, deliberate, model-based planning allows the assessment of multiple courses of action and their potential consequences. However, the computations involved in planning take a long time and might not be the best solution under time-pressure conditions. In addition, it requires that a model of the environment has been formed in the first place. On the other hand, habitual or model-free decision-making systems are much faster and therefore more suitable for contexts in which fast decisions need to be made. They do not rely on the acquisition of a model; however, they take much longer to be learned \cite{Daw2018,Blundell2016}. Besides these two modes of operation, some researchers have argued that episodic control represents a third type of system that could also play a role in generating adaptive behavior \cite{Lengyel}. An episodic controller operates by remembering the action that led to the best outcome in a given situation. It is computationally lighter than model-based algorithms and does not suffer from too much uncertainty or noise due to the complex calculations involved in forward search. Moreover, it takes much less time to acquire than a habit. Therefore, in situations when an animal is exploring a novel environment, and no model, policy, or habit has yet been formed, relying on the fast, instance-based learning provided by episodic control systems like SEC might be of critical importance.

%% ORCHESTRATION BETWEEN DIFFERENT SYSTEMS - OPTIMAL AT DIFF RANGES
In the framework proposed by Lengyel and Daw, each of these learning systems (model-based/deliberate, model-free/habitual, and instance-based/episodic) has its trade-offs, derived from their intrinsic properties and inductive biases \cite{Lengyel}. By their very nature, these control systems have an optimal performance at different stages of the learning process of the agent and could be operating at different timescales \cite{Botvinick2019} as also demonstrated by the relation between the reactive, adaptive and contextual layers of the DAC framework \cite{Verschure2014}. Understanding the regimes in which they optimally operate and how they combine to attain adaptive behavior in physical and social environments is of paramount importance if we aim to build synthetic embodied cognitive systems able to exhibit distributed adaptive control in complex worlds.

\section*{Data availability}
The data that support the findings of this study are available in Zenodo with the identifier(s): 10.5281/zenodo.11506323

\section*{Code availability}
The custom code use to generate the findings of this study are available in a code repository with the identifier(s): https://github.com/IsmaelTito/SEC

\section*{Acknowledgments}
This study was funded by the Counterfactual Assessment and Valuation for Awareness Architecture (CAVAA) project (European Innovation Council's Horizon program, grant ID: 101071178).

\bibliographystyle{elsarticle-num} 
\bibliography{bibliography_v2}





% \end{thebibliography}

\appendix
\section{Sequential Episodic Control Hyperparameters} 
\label{sec:appendix1:hyperparameters}

Hyperparameter values of both SEC and NSEC algorithms for the AnimalAI double T-maze task. 

\begin{center}
\begin{tabular}{||c | c | c ||} 
 \hline
 Hyperparameter & Value & Description  \\ [0.5ex] 
 \hline\hline
 Episodes & 5000 & Episodes performed by each agent \\ 
 \hline
 n & 50 & STM buffer length  \\
 \hline
 m & 500 & LTM length \\
 \hline
 State vector & 20 & Compressed feature representations length \\
 \hline
 $\alpha$ & 0.1 & Increase of the sequential bias term  \\
  \hline
 $\beta$ & 0.05 & Decay of the sequential bias term  \\
 \hline
 $\tau$ & 0.9 & Discount factor  \\ 
 \hline
 $\theta_{prop}$ & 0.98 & Relative threshold   \\ 
 \hline
 $\theta_{abs}$ & 0.995 & Absolute threshold  \\ [1ex] 
 \hline
\end{tabular}
\end{center}

%%%%%%

\section{Convolutional Autoencoder} 
\label{sec:appendix2:autoencoder}

The architecture of the convolutional autoencoder used for the embedding function. Implemented in Keras. The optimizer used is RMSprop, with a learning rate of 0.001.  

\begin{center}
\begin{tabular}{||c | c | c | c | c | c ||} 

 \hline
 Number & Type & Units & Kernel & Padding & Activation  \\ [0.5ex] 
 \hline\hline
 \multicolumn{6}{||c||}{Encoder} \\ 
 \hline
 1 & Conv2D & 16 & (3,3) & same & ReLU  \\
 2 & MaxPooling2D &  & (2,2) & same &   \\
 3 & Conv2D & 32 & (3,3) & same & ReLU  \\
 4 & MaxPooling2D &  & (2,2) & same &   \\
 5 & Conv2D & 32 & (3,3) & same & ReLU  \\
 6 & MaxPooling2D &  & (2,2) & same &   \\
 7 & Flatten &  &  &  &   \\
 8 & Dense & 20 &  &  & ReLU  \\
 \hline
 \multicolumn{6}{||c||}{Decoder} \\ 
 \hline
 9 & Dense & 11*11*32 &  &  &  \\
 10 & Reshape & (11,11,32) &  &  &   \\
 11 & Conv2D & 32 & (3,3) & same & ReLU  \\
 12 & UpSampling2D &  & (2,2) &  &   \\
 13 & Conv2D & 32 & (3,3) & same & ReLU  \\
 14 & UpSampling2D &  & (2,2) &  &   \\
 15 & Conv2D & 16 & (3,3) & same & ReLU  \\
 16 & UpSampling2D &  & (2,2) &  &   \\
 17 & Conv2D & 3 & (3,3) & same & ReLU  \\
 \hline
  
\end{tabular}
\end{center}

\end{document}